%% file: main.tex
\documentclass[letterpaper, 10 pt, conference]{ieeeconf}  

\IEEEoverridecommandlockouts                              

\usepackage{cite}
\usepackage{fancyhdr, graphics, graphicx} 
\usepackage{epsfig} 
\usepackage{mathptmx} 
\usepackage{times} 
\usepackage{amsmath} 
\usepackage{amssymb}  
\usepackage{amsfonts}
\usepackage[ruled,vlined]{algorithm2e}
\usepackage{algorithmic}
\usepackage{bm}
\usepackage{booktabs}
\usepackage{dcolumn,tipa}
\newcolumntype{d}[1]{D{.}{.}{#1}}
\usepackage{etoolbox}
\AtBeginEnvironment{tcolorbox}{\small}
\usepackage{fixltx2e}
\usepackage{gensymb}
\usepackage{hyperref}
\usepackage{listings}
\usepackage{makecell}
\usepackage{multirow}
\usepackage{pifont}
\usepackage{tabularx}
\usepackage[most]{tcolorbox}
\usepackage{textcomp}
\usepackage{todonotes}
\usepackage{xcolor}
\usepackage{cleveref}
\usepackage{caption}
\usepackage{subcaption}

\usepackage{xspace}
\newcommand{\etal}{\textit{et~al}.\xspace}
\tcbuselibrary{breakable}
\usepackage[activate={true,nocompatibility},final,tracking=true,kerning=true,spacing=true,factor=1100,stretch=10,shrink=10]{microtype}
\microtypesetup{protrusion=false}
\newcommand*\colorcheck[1]{%
  \expandafter\newcommand\csname #1check\endcsname{\textcolor{#1}{\ding{51}}}%
}
\newcommand*\colorcross[1]{%
  \expandafter\newcommand\csname #1cross\endcsname{\textcolor{#1}{\ding{55}}}%
}
\colorcheck{green}
\colorcross{red}

\usepackage{marginnote}

\renewcommand{\baselinestretch}{0.98}
\input{PDDL}

\title{\LARGE \bf
DELTA: Decomposed Efficient Long-Term Robot Task Planning\\using Large Language Models
}

\author{Yuchen Liu$^{1, 2}$, Luigi Palmieri$^{1}$, Sebastian Koch$^{1}$, Ilche Georgievski$^{2}$ and Marco Aiello$^{2}$
\thanks{$^{1}$Corporate Research and Advance Engineering, Robert Bosch GmbH, Germany
        {\tt\small \{yuchen.liu2, luigi.palmieri, sebastian.koch2\}@bosch.com}}%
\thanks{$^{2}$Institute of Architecture of Application Systems, University of Stuttgart, Germany
        {\tt\small \{yuchen.liu, ilche.georgievski, marco.aiello\}@iaas.uni-stuttgart.de}}%
\thanks{This work was partly supported by the EU Horizon 2020 research and innovation program under grant agreement No. 101017274 (DARKO).}
}

\begin{document}
\bstctlcite{IEEEexample:BSTcontrol}
\include{pythonlisting}

\maketitle
\thispagestyle{empty}
\pagestyle{empty}

\begin{abstract}

Recent advancements in Large Language Models (LLMs) have sparked a revolution across many research fields.
In robotics, the integration of common-sense knowledge from LLMs into task and motion planning has drastically advanced the field by unlocking unprecedented levels of context awareness. 
Despite their vast collection of knowledge, large language models may generate infeasible plans due to hallucinations or missing domain information. To address these challenges and improve plan feasibility and computational efficiency, we introduce DELTA, a novel LLM-informed task planning approach. 
By using scene graphs as environment representations within LLMs, DELTA achieves rapid generation of precise planning problem descriptions. To enhance planning performance, DELTA decomposes long-term task goals with LLMs into an autoregressive sequence of sub-goals, enabling automated task planners to efficiently solve complex problems.  
In our extensive evaluation, we show that DELTA enables an efficient and fully automatic task planning pipeline, achieving higher planning success rates and significantly shorter planning times compared to the state of the art.
Project webpage: \url{https://delta-llm.github.io/}

\end{abstract}

\section{INTRODUCTION}
\begin{figure}[t]
    \centering
    \includegraphics[width=0.475\textwidth]{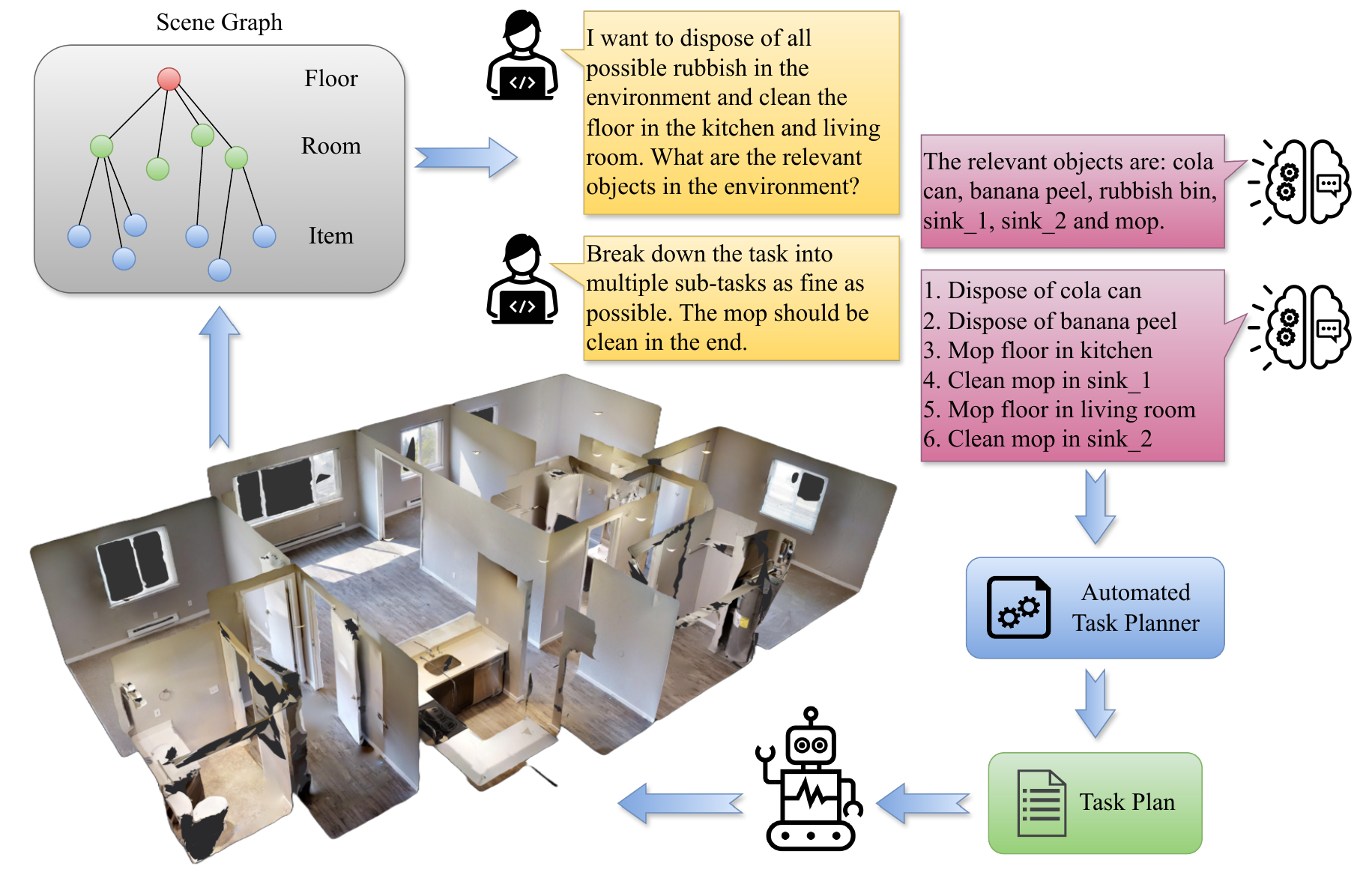}
    \caption{An example of the long-term task decomposition. A Scene Graph (SG) is pre-built from the environment~\cite{Armeni2019hierachical}. Using the SG as the environment representation, a human user queries a LLM with goal descriptions to extract the relevant items and decompose the goal into multiple sub-goals. An automated task planner generates a task plan with respect to the sub-goals for the robot to execute.\vspace{-1em}}
    \label{fig:fig1}
\end{figure}

With the rapid and enormous progress in the research field of Natural Language Processing (NLP), various powerful Large Language Models (LLMs) have been developed that are capable of producing human-like texts, programming code, and service compositions etc.~\cite{openai2023gpt4, chowdhery2023palm, devlin2018bert, touvron2023llama, aiello2023service}. 
Nowadays with more and more robots cooperating with humans in industrial and household settings~\cite{liu2023human, aboki2023automating}, 
e.g., performing household tasks such as cleaning (Fig. \ref{fig:fig1}), many researchers use LLMs for solving robot Task And Motion Planning (TAMP) problems~\cite{huang2022language, song2023llm, liu2023llm+, ding2023task, ahn2022i, rana2023sayplan, chen2023autotamp, silver2024generalized}. While directly using pre-trained LLMs to generate action plans for the robots tends to result in extremely low success rates in generating executable plans and completing the goals~\cite{pallagani2024prospects, valmeekam2022large}, most of them use LLMs to extract common-sense knowledge to improve the performance of classical automated task planning approaches with respect to plan correctness, executability, and feasibility~\cite{huang2022language, ding2023task, chen2023autotamp}. Several approaches use LLMs to generate task specifications defined in formal language, e.g., the domain and problem files programmed in the Planning Domain Definition Language (PDDL)~\cite{mcdermott1998pddl},
that can be solved by the off-the-shelf TAMP algorithms~\cite{liu2023llm+, xie2023translating, zuo2024planetarium}. 
However, previous TAMP approaches were cumbersome as they required vast manual knowledge engineering and input from human experts, inducing practically impossible domain adaptations.
On the other hand, none of the approaches above tackle long-term task planning problems, which are particularly difficult to solve with the growing problem complexity \cite{chen2023autotamp}.

For robots solving long-term task sequences in large and complex environments, having efficient environment representations is a crucial prerequisite for the robot to understand the semantic information~\cite{goel2023semantically}.
While mapping the mid-level perceptual representations (e.g., 2D semantic segmentation) into more appropriate high-level abstractions (e.g., environment topology and semantic relations between objects) can be costly and complex~\cite{chalvatzaki2023learning},
it can be solved efficiently using high-level representations such as Scene Graphs (SGs)~\cite{ravichandran2022hierarchical}. For tackling task planning problems in such environments, researchers have found that SGs can serve as compact and informative spatial representations and can improve planning efficiency~\cite{agia2022taskography, chalvatzaki2023learning, rana2023sayplan}.

As it emerges from the state of the art, utilizing LLMs and automated task planning techniques to solve long-term robot task planning problems, with structured representations of large environments, still remains an open research topic. 
Therefore, we propose \textbf{DELTA}: \emph{\textbf{D}ecomposed \textbf{E}fficient \textbf{L}ong-term \textbf{TA}sk planning for mobile robots using LLMs}, which is the first, to the best of our knowledge, to fill the aforementioned vacancy.
DELTA first feeds SGs into LLMs to generate the necessary domain and problem specifications in formal planning language, then decomposes the long-term task goals into multiple sub-ones using LLMs. The corresponding sub-problems are then solved autoregressively with an automated task planner.
In summary, we present the following key contributions:

\emph{i)} We introduce a novel combination of LLMs and SGs that enables the extraction of actionable and semantic knowledge from LLMs and its grounding into the environmental topology. Thanks to one-shot prompting, DELTA is capable of solving complex planning problems in unseen domains.

\emph{ii)} We show that with the LLM-driven task decomposition strategy and the usage of formal planning language, compared to representative LLM-based baselines, DELTA is able to complete long-term tasks with higher success rates, near-optimal plan quality, and significantly shorter planning time. 

\section{RELATED WORK}\label{RW}

\subsection{3D Scene Graphs}
A 3D Scene Graph (3DSG) is a recent 3D scene representation used to model large real-world environments as a graph structure. 
They were first introduced by Armeni \etal \cite{Armeni2019hierachical} as a hierarchical model to connect buildings, rooms, objects, and humans in multiple layers.  Following this introduction, Rosinol \etal \cite{rosinol20203d,Rosinol_2021_SAGE} and Hughes \etal \cite{hughes2022hydra} investigated the construction of 3DSGs from sensor data. 
While other approaches focus on modeling semantic relationships between objects \cite{Wald2020D3SSG, Wu_2021_sgfusion, Wang_2023_vlsat, koch2024sgrec, koch2023lang3dsg} from 3D point clouds.
With the rise of LLMs, 3DSGs also become open-vocabulary \cite{conceptgraphs,chang2023contextaware,koch2024open3dsg}, offering a deeper understanding of relationships between objects.
Recently, 3DSGs have started to be integrated into robotics systems. Applications include navigation \cite{seymour2022graphmapper,conceptgraphs,chang2023contextaware} as well as task and motion planning \cite{agia2022taskography,rana2023sayplan}. 

\subsection{LLM-based Robot Task and Motion Planning}
The rich embedded semantic and common-sense knowledge allows LLMs to proficiently understand Natural Language (NL) instructions and perform temporal reasoning, first-order logic translation, and few-shot or even zero-shot planning \cite{xiong2024large, yang2023harnessing, brown2020language}. However, 
LLMs still struggle with analyzing complex spatial relationships and processing detailed environmental features \cite{pallagani2024prospects}. Consequently, plans directly generated by LLMs are often not executable for the robots. 
Therefore, researchers have developed various approaches to ground LLM's output into executable and affordable action plans, or ground actionable knowledge into formal planning or programming language.
Ahn \etal \cite{ahn2022i} proposed \textit{SayCan} to constrain the LLMs with pre-trained skills when generating actions.
Liu \etal \cite{liu2023llm+} introduced \textit{LLM+P} that translates NL problem descriptions into PDDL problem files with LLMs given user-provided PDDL domain files. Silver \etal \cite{silver2024generalized} leveraged LLMs to comprehend the domain knowledge and problem specifications from PDDL files, then used LLMs to generate Python code to solve the problems.
However, most of these approaches require handcrafted domain descriptions provided by human experts, and do not generalize to new domain knowledge.
Moreover, they do not tackle long-term planning problems in large and complex environments.

Semantic understanding is a crucial factor for robot navigation in large environments. LLMs unlock the capability of reasoning over semantic relations embedded in large scenes and allow the capture of those relations from different scene representations, e.g., semantic maps \cite{chen2023open}, landmarks \cite{shah2023lm, wang2022less}, as well as SGs.
Rana \etal \cite{rana2023sayplan} proposed \textit{SayPlan} that uses LLMs to first conduct a semantic search through 3DSGs, then generates task plans upon the graph and refines iteratively, achieving grounded and scalable robot TAMP. But it still does not aim to solve long-term tasks.

However, since the probability of LLMs in producing incorrect output accumulates with growing planning horizon \cite{pallagani2024prospects}, most of the LLM-based approaches above have difficulties in tackling long-term planning problems. Thus, they mainly focus on semantically simple short-term tasks, e.g., object rearrangement, object-goal navigation, or other tasks that consist of a few such sub-tasks. 
The capability of LLMs to handle long-term tasks in large environments is not fully exploited.
While decomposing a long-term task into multiple sub-tasks via classical machine learning methods can lead to a significant reduction of planning time \cite{liu2023human}, completing such a job with LLMs is still unexplored in the state-of-the-art.

\begin{figure}[t]
    \centering
    \includegraphics[width=0.25\textwidth]{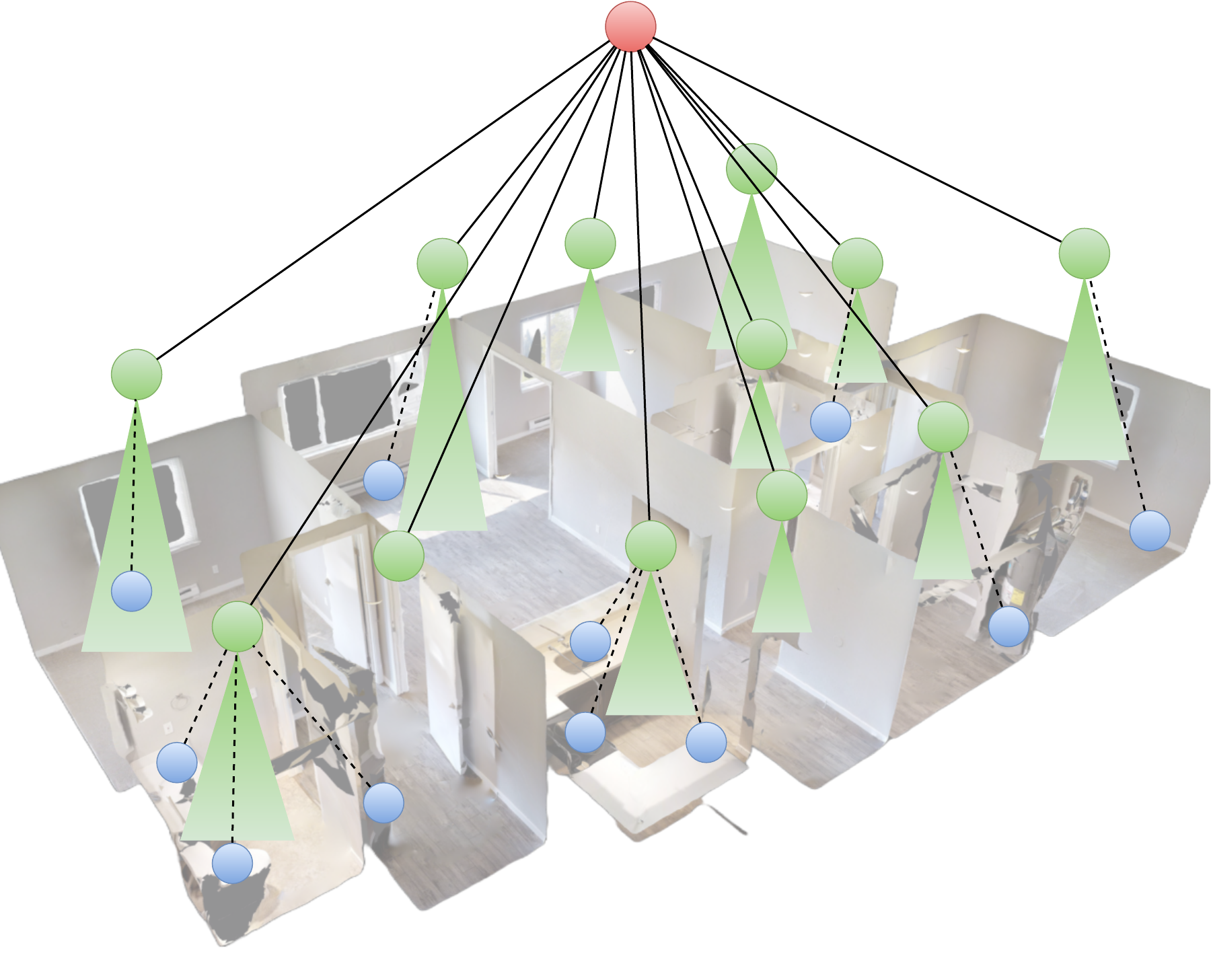}
    \caption{\textit{Shelbiana} scene \cite{Armeni2019hierachical} and the corresponding SG with \textcolor{red}{floor}, \textcolor{green}{room}, and \textcolor{blue}{item} node layers. The edges refer to the semantic relationships. Not all item nodes are visualized.\vspace{-1em}}
    \label{fig:sg}
\end{figure}

\section{METHODOLOGY}\label{algo}

\begin{figure*}[t]
    \centering
    \includegraphics[width=0.9\textwidth]{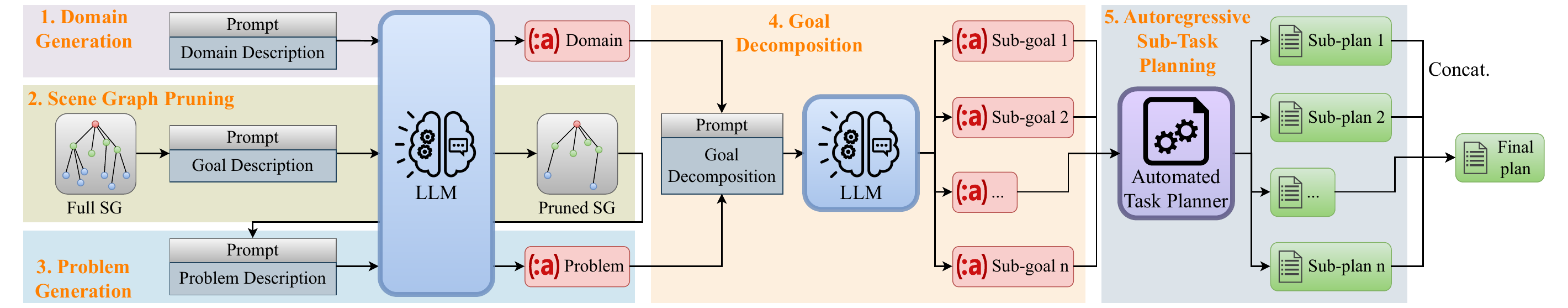}
    \caption{The system architecture of DELTA with five steps: \textit{Domain Generation}, \textit{Scene Graph Pruning}, \textit{Problem Generation}, \textit{Goal Decomposition}, and \textit{Autoregressive Sub-Task Planning}.\vspace{-2em}}
    \label{fig:sys}
\end{figure*}

\subsection{Problem Statement}
We focus on solving long-term robot task planning problems with LLMs and consider mobile robot navigation tasks in household environments. The approach can also be generalized to other use cases.
Given a SG as environment representation and domain and problem descriptions in NL, the LLM will generate the PDDL planning files and decompose the long-term goal into a sequence of sub-goals for solving the corresponding sub-problems autoregressively.

To distinguish the \textit{object} keyword in PDDL from objects in SGs, in the following, we refer to the objects in SGs as \textit{items}. We define \textit{agent} as an object type in the domain file and \textit{robot} as an instance of \textit{agent} in the problem file. Furthermore, we assume full observability of all nodes in the SGs.

\subsection{System Architecture}

The architecture of DELTA is built around a five-step process (Fig.~\ref{fig:sys}): domain generation, scene graph pruning, problem generation, goal decomposition, and autoregressive sub-task planning.

\begin{lstlisting}[
  float=t,
  basicstyle=\fontsize{8}{8}\selectfont\ttfamily,
  caption={Action ``\textit{mop floor}'' defined in PDDL},
  label={lst:pddl_act},
  language=PDDL,
  belowskip=-2\baselineskip]
(:action mop_floor
    :parameters (?a - agent ?i - item ?r - room)
    :precondition (and
        (agent_at ?a ?r)
        (item_is_mop ?i)
        (item_pickable ?i)
        (agent_has_item ?a ?i)
        (mop_clean ?i)
        (not(floor_clean ?r))
    )
    :effect (and
        (floor_clean ?r)
        (not(mop_clean ?i))
        (not(battery_full ?a))
    )
)
\end{lstlisting}

\subsubsection{Domain Generation}
The LLM takes an NL prompt describing the domain knowledge as input and generates a domain description file encoded in formal planning language, e.g., PDDL, correspondingly in a one-shot fashion. The prompt consists of three main parts: role, example, and instruction. The further prompts in the following steps also have the same structure.
The example of the domain description includes the necessary object types and the action knowledge, i.e., pre-conditions and effects. For instance, the ``\textit{mop\_floor}'' action can be described as ``\textit{For mopping the floor, 
the agent is in the room and has the mop in hand, the mop is clean while the floor is not clean. After the action, the floor is clean, but the mop is not clean anymore, and the agent's battery will no longer be full.}'' The corresponding action can be formulated in PDDL as shown in Listing \ref{lst:pddl_act}.\looseness=-1

Subsequently, the instruction introduces the requirements for generating a new domain file.
An overview of the prompt structure is shown as follows (the \textcolor{purple}{\textit{purple}} and \textcolor{blue}{\textit{blue}} text refers to NL description and programming code, respectively):

\begin{tcolorbox}[breakable, colback=black!5!white, colframe=black!75!black, left=2pt, right=2pt, top=2pt, bottom=2pt]
\textbf{Role}: You are an excellent domain generator. Given a description of domain knowledge, you can generate a PDDL domain file.

\textbf{Example}: A robot in a household environment can perform the following \textcolor{purple}{\textit{example object types}} and \textcolor{purple}{\textit{example actions with pre-conditions and effects}}.
The corresponding action definitions in a PDDL domain file look like: \textcolor{blue}{\textit{example\_domain.pddl}}.

\textbf{Instruction}: A new domain has the following \textcolor{purple}{\textit{new object types and actions}}. Please generate a corresponding PDDL domain file.
\end{tcolorbox}


\subsubsection{Scene Graph Pruning}
SG has a hierarchical structure,
An example of the layout is shown in Fig.~\ref{fig:sg}. The room nodes are annotated with their neighboring rooms, and the item nodes contain several attributes, e.g., accessibility, states, and affordable actions.
In particularly large environments, SGs can contain a large number of items, where not all items are relevant for accomplishing different tasks. 
Therefore, we prune the SGs with LLMs in the following way:

\begin{tcolorbox}[breakable, colback=black!5!white, colframe=black!75!black, left=2pt, right=2pt, top=2pt, bottom=2pt]
\textbf{Role}: You are an excellent assistant in pruning SGs with a list of SG items and a goal description.

\textbf{Example}: A SG can be programmed as a nested Python dictionary such as \textcolor{blue}{\textit{example\_sg.py}}. For accomplishing the \textcolor{purple}{\textit{example goal}}, the relevant items are \textcolor{blue}{\textit{[example\_relevant\_items]}}.

\textbf{Instruction}: Given a new \textcolor{blue}{\textit{query\_sg.py}} and a new \textcolor{purple}{\textit{goal description}}, please prune the SG by keeping the relevant items.
\end{tcolorbox}


Pruning the SG allows a reduction of input tokens for the LLMs, thus achieving a fast response time in generating the problem files. On the other hand, the more concise the information provided to the LLMs, the less likely that the LLMs generate erroneous output and hallucinations \cite{pallagani2024prospects}.

\begin{lstlisting}[
  float=t,
  basicstyle=\fontsize{8}{8}\selectfont\ttfamily,
  caption={Goal states of the house cleaning problem in PDDL},
  label={lst:pddl_goal},
  language=PDDL,
  belowskip=-1.5\baselineskip]
(:goal
    (and
        (item_disposed cola_can)
        (item_disposed banana_peel)
        (floor_clean living_room)
        (floor_clean kitchen)
        (mop_clean mop)
    )
)
\end{lstlisting}

\subsubsection{Problem Generation}
The prompt of this step can be similarly formulated in the following way:

\begin{tcolorbox}[breakable, colback=black!5!white, colframe=black!75!black, left=2pt, right=2pt, top=2pt, bottom=2pt]
\textbf{Role}: You are an excellent problem generator. Given a SG and desired goals, you can generate a PDDL problem file.

\textbf{Example}: Given an \textcolor{blue}{\textit{example\_sg.py}}, an \textcolor{purple}{\textit{example goal description}}, and using the predicates defined in \textcolor{blue}{\textit{example\_domain.pddl}}, a corresponding PDDL problem file looks like: \textcolor{blue}{\textit{example\_problem.pddl}}.

\textbf{Instruction}: Given a new \textcolor{blue}{\textit{query\_sg.py}}, a new \textcolor{purple}{\textit{goal description}}, please generate a new PDDL problem file using the predicates in the previously generated \textcolor{blue}{\textit{query\_domain.pddl}}. 
\end{tcolorbox}

In the generated problem file, the connections of rooms in the SG can be expressed using the \textit{neighbor} predicate. For instance, if \textit{kitchen} is connected with \textit{corridor}, the relationship can be formulated as \textit{(neighbor kitchen corridor)} and \textit{(neighbor corridor kitchen)} since the rooms are connected bi-directionally. 
Similarly, the attributes of items can be defined with the predicates from the previously generated domain file, such as their positions and accessibilities.
The LLM also translates the NL goal description into PDDL. E.g., the goal of a house cleaning problem given by the human user can be formulated as shown in Listing~\ref{lst:pddl_goal}, Fig. \ref{fig:fig1}.

\subsubsection{Goal Decomposition}
To improve the computational efficiency and reduce the complexity of the planning problem, the long-term goal defined in the problem file can be decomposed with LLMs using the following prompt: 

\begin{tcolorbox}[breakable, colback=black!5!white, colframe=black!75!black, left=2pt, right=2pt, top=2pt, bottom=2pt]
\textbf{Role}: You are an excellent assistant in decomposing long-term goals. Given a PDDL problem file, you can decompose the goal states into a sequence of sub-goals.

\textbf{Example}: Given an \textcolor{blue}{\textit{example\_problem.pddl}}, the goal states can be decomposed into a sequence of \textcolor{purple}{\textit{example sub-goals}}. Using the predicates defined in \textcolor{blue}{\textit{example\_domain.pddl}}, the \textcolor{purple}{\textit{example sub-goals}} can be formulated as: \textcolor{blue}{\textit{sub-goal\_1.pddl}}, ..., \textcolor{blue}{\textit{sub-goal\_n.pddl}}.

\textbf{Instruction}: Given the \textcolor{blue}{\textit{query\_problem.pddl}} generated previously, please decompose the goal considering the predicates and actions from the previously generated \textcolor{blue}{\textit{query\_domain.pddl}}.
\end{tcolorbox}

Taking action knowledge into account in goal decomposition is essential. For instance, knowing that the action \textit{mop\_floor} requires \textit{mop\_clean} and results in \textit{not (mop\_clean)} as shown in Listing \ref{lst:pddl_act}, which infers that one cannot mop the floor in another room continuously since the mop turns dirty after mopping the previous room. Thus, cleaning the mop before mopping the next room should be considered when decomposing the goal, as shown in Fig. \ref{fig:fig1}.

\subsubsection{Autoregressive Sub-Task Planning}
We use an automated task planner to solve the corresponding sub-problems one after another as shown in Algorithm~\ref{alg:iter_plan}. It takes the planner $\Pi$, the previously generated PDDL domain file $d$ and problem file $p_0$ with undecomposed goals, and the sequence of PDDL sub-goals $\bm{G}$ as inputs. 
The initial states $\bm{s}_1$ of the first sub-problem $p_1$ are identical to the initial states $\bm{s}_0$ from the original problem $p_0$. Thus, $p_1$ can be simply formulated by replacing the undecomposed goal states $\bm{g}_0$ from $p_0$ with the first sub-goal states $\bm{g}_1$ from $\bm{G}$.

The final states resulting from the plan's execution of each solvable sub-problem are, in fact, exactly the initial states of the next sub-problem. Therefore, as shown in the for-loop in L. \ref{alg:for_start}-\ref{alg:for_end}, after solving each sub-problem $p$, we obtain a sub-plan $\pi'$ and the resulting final states $\bm{s'}$ (L. \ref{alg:plan}), which will then be assigned to $\bm{s}$ for the next sub-problem.
The following sub-problems can be solved in the same way autoregressively. The final task plan $\pi$ can be obtained by concatenating all sub-plans $\pi'$ (L. \ref{alg:concat}), that only consist of executable actions.

\section{EVALUATION}\label{experiment}
In this section, we detail the metrics, domains, baselines, and datasets used for the evaluation.
\subsection{Metrics}
We evaluate the proposed system in terms of computational and task efficiency by the following metrics:
\begin{itemize}
    \item \textbf{Success rate}: ratio of the succeeded trials to all trials. A trial is successful if the plan validator reports that the generated plan is valid, i.e., correct and executable. The success rates of each domain are averaged through the experiments with all three scenes.
    \item \textbf{Plan length}: number of actions in a plan. The decomposed plan length shows the length of the concatenated sub-plans from solving the sub-problems.
    \item \textbf{Planning time}: inference time of the automated planner for finding a solution. The decomposed planning time refers to the total time of solving all sub-problems.
    \item \textbf{Number of expanded nodes}: nodes in a search tree created and explored during the search process for solving a planning problem. A lower number implies lower problem complexity reported by the planner.
\end{itemize}

\subsection{Evaluation Domains}
We run the evaluation on five domains. The \textbf{Laundry} domain has a short-term task that serves as the example for one-shot prompting, where the robot is asked to bring the dirty clothes and detergent to the washing machine and then bring them to the bedroom after washed.
Of the other domains, two have \textbf{independent} sub-tasks, namely, 
the \textbf{PC Assembly} domain (in the following abbreviated as \textit{PC}) requires the robot to gather six different PC parts,
i.e., a mainboard, a CPU, a GPU, a RAM, a SSD, and a Power Supply Unit (PSU), 
distributed in the environment and bring them to the living room for assembly. 
In the \textbf{Dining Table Setup} domain (abbr. \textit{Dining}), the robot should collect a plate, a fork, a knife, a spoon, and a glass from different rooms, and also find something romantic, then place them on the dining table. Both domains can be decomposed into independent transportation sub-tasks. 
The remaining two domains have \textbf{dependent} sub-tasks which should be executed in a certain order.
In the \textbf{House Cleaning} domain (abbr. \textit{Cleaning}), the robot should first dispose of a cola can, a banana peel, and a rotting apple in the rubbish bin, then mop the floor in the kitchen and living room and clean the mop immediately after cleaning each room. Finally, the robot should return to the hub for recharging. The \textbf{Home Office Setup} domain (abbr. \textit{Office}) requires the robot to set up a home office in the living room by bringing a desk, a lamp, a shelf, and a locker. The shelf and locker have contents inside that should be kept in the end, but they cannot be moved without unloading the contents. 
In each domain, the robot can only load one item at a time.

\setlength{\textfloatsep}{4pt}
\begin{algorithm}[t]
\small
\DontPrintSemicolon
\LinesNumbered
\SetAlgoLined
\KwData{$\Pi$, $d$, $p_0$, $\bm{G}$}
\KwResult{$\pi$}
$\pi \leftarrow \varnothing$\; \label{alg:init}
extract $\bm{s}_0, \bm{g}_0$ from $p_0$\;
$\bm{s} \leftarrow \bm{s}_0$\;
\For{$\bm{g} \in \bm{G}$}{ \label{alg:for_start}
    $p \leftarrow$ replace $\bm{s}_0, \bm{g}_0$ in $p_0$ with $\bm{s}, \bm{g}$\; \label{alg:replace}
    $\pi', \bm{s'} \leftarrow \Pi(d, p)$\; \label{alg:plan}
    $\pi \leftarrow$ concat$(\pi, \pi')$\; \label{alg:concat}
    $\bm{s} \leftarrow \bm{s'}$\;
} \label{alg:for_end}
\caption{Autoregressive Sub-Task Planning}
\label{alg:iter_plan}
\end{algorithm}

\subsection{Baselines} \label{sec:baseline}
We select four most popular and representative LLM-based task planning approaches: LLM-As-Planner, LLM+P \cite{liu2023llm+}, LLM-GenPlan \cite{silver2024generalized}, and SayPlan \cite{rana2023sayplan} according to the usage of formal language and environmental representation, as well as the capability of tackling long-term tasks shown in Table~\ref{tab:cap}. 

\textbf{LLM-As-Planner} is a naive approach that directly queries the LLM to generate a high-level plan using a prompt that comprises all information, i.e., domain knowledge, environment and goal descriptions.

\textbf{LLM+P} \cite{liu2023llm+} uses LLMs to translate NL problem descriptions into a PDDL problem file given user-provided PDDL domain files,
for an automated task planner to solve the problem. It can be treated as a subset of DELTA with only the problem generation step.
In the following experiments, we provide a pre-defined PDDL domain file as input.

\textbf{LLM-GenPlan} \cite{silver2024generalized} uses LLMs to first summarize the domain knowledge from input PDDL files, then propose a simple generalized planning strategy without using search, and finally generates Python code that outputs a task plan. The LLMs can refine the code using the debug information from a plan validator with maximal $4$ iterations.

\textbf{SayPlan} \cite{rana2023sayplan} first determines a task-relevant sub-SG with the LLM-based semantic search, then uses LLMs to generate a high-level plan using the sub-SG and iteratively replans based on environmental feedback. We implemented SayPlan on our own due to the lack of available open-source code.
The number of maximal replanning iterations is set to $4$.

\subsection{Dataset}
We use four scene graphs from the 3D Scene Graph dataset \cite{Armeni2019hierachical}: \textit{Kemblesville} ($9$ rooms, $16$ items) is paired with the \textit{Laundry} domain. \textit{Allensville} ($11$ rooms and $42$ items), \textit{Parole} ($7$ rooms, $31$ items), and \textit{Shelbiana} ($12$ rooms, $34$ items) are used to evaluate the rest four domains. 
We implement the SGs as nested dictionaries in Python. 

\begin{table}[t]
    \centering
    \renewcommand{\arraystretch}{0.92}
    \begin{tabular}{l *{3}{c}}
        \toprule
        \text{Models} & \makecell{Formal planning\\language} & \makecell{Structured\\env. repr.} & \makecell{Long-term\\tasks}\\
        \midrule
        \text{LLM-As-Planner} & \redcross & \redcross & \redcross \\
        \text{LLM+P} \cite{liu2023llm+} & \greencheck & \redcross & \greencheck \\
        \text{LLM-GenPlan} \cite{silver2024generalized} & \greencheck & \redcross & \redcross \\
        \text{SayPlan} \cite{rana2023sayplan} & \redcross & \greencheck & \redcross \\
        \text{DELTA (ours)} & \greencheck & \greencheck & \greencheck \\
        \bottomrule
    \end{tabular}
    \caption{Capabilities of different LLM-based models}
    \label{tab:cap}
\end{table}

\subsection{Implementation and Parameters}
We evaluate DELTA with pre-trained \textit{GPT-4-turbo} (version 2024-04-09), \textit{GPT-4o} (version 2024-05-13), and \textit{Llama-3.1-70B} with default \textit{temperature} and \textit{top\_p} parameters.
The other baselines are evaluated with \textit{GPT-4o}. 

We use Fast Downward (FD) \cite{helmert2006fast} automated task planner with the default search configuration \textit{seq-opt-lmcut} and the timeout of $60 s$.
Moreover, we use PDDLGym \cite{silver2020pddlgym} to obtain the world states, and the plan validation tool VAL \cite{howey2004val} to validate the correctness and executability of the generated plans. Each experiment is repeated with $50$ trials, resulting in $600$ trials crosswise evaluated with $4$ domains and $3$ scenes in total. All approaches are implemented in Python $3.8.13$. We run the experiments on a standard PC with an Intel~Xeon~W CPU at $3.40$ GHz and $32$ GB RAM. 
The \textit{GPT} models are deployed with Azure OpenAI Service, while the \textit{Llama} model with two Nvidia A100 GPUs in $4$-bit quantization.

\section{RESULTS AND DISCUSSION}
The evaluation results are displayed in Tables~\ref{tab:sr} and~\ref{tab:result_all}.
\textbf{LLM-As-Planner} performs the worst among all approaches with $70\%$ and $38.67\%$ in domains with independent sub-tasks, and no successful case in those with dependent sub-tasks. 
Despite the ability in temporal reasoning, LLMs still have difficulties in discovering the underlying dependencies and preconditions of complex long-term tasks \cite{xiong2024large, valmeekam2022large}.

By grounding NL into formal planning language, \textbf{LLM+P} is able to achieve slightly higher success rates in the \textit{PC} domain and it is also able to reach the optimal plan length and number of expanded nodes. 
However, it only reaches $4\%$ success rate in \textit{Dining} and never succeeds in \textit{Cleaning} and \textit{Office} domains. The leading failure is planner timeout.
Since LLM+P consumes the original SGs with a large number of items, although the LLMs have mostly transferred the items from SGs to the PDDL problem files, the complexity of the planning problem increases exponentially with the growing number of items, resulting in exceeding the planner's timeout.

\begin{table}[t]
    \centering
    \renewcommand{\arraystretch}{0.95}
    \begin{tabular}{l *{4}{c}}
        \toprule
        Models & PC & Dining & Cleaning & Office \\
        \midrule
        LLM-As-Planner           & 70 & 38.67 & 0 & 0 \\
        LLM+P                    & 76 & 4 & 0 & 0 \\
        LLM-GenPlan (w/o rp.)    & 36.67 & 38.67 & 0 & 0 \\
        LLM-GenPlan              & 88 & 80.67 & 3.33 & 0.67 \\
        SayPlan (w/o rp.)        & 8.67 & 1.33 & 0 & 0.67 \\
        SayPlan                  & 68.67 & 70.67 & 54 & 40 \\
        \midrule
        DELTA (Llama-3.1-70B)    & 34.67 & 23.33 & 0 & 0.67 \\
        DELTA (GPT-4-turbo)      & 93.33 & 74 & 32.67 & 9.33 \\
        DELTA (GPT-4o, w/o dp.)  & 97.33 & 99.33 & \textbf{80} & 68.67 \\
        DELTA (GPT-4o)           & \textbf{98} & \textbf{100}   & \textbf{80} & \textbf{74.67} \\
        \bottomrule
    \end{tabular}
    \caption{Success rates $[\%]$ of different models. The results from each domain are averaged through all the scenes. The upper part of the table shows the results of the baselines based on \textit{GPT-4o}. The lower part lists the outcome of DELTA. \textit{w/o rp.} and \textit{w/o dp.} refer to without replanning and goal decomposition, respectively.}
    \label{tab:sr}
\end{table}


\textbf{LLM-GenPlan} learns the domain knowledge encoded in PDDL and generalizes to solve unseen tasks. It achieves around $80\%$ success rates and near-optimal plan lengths in the domains with independent sub-tasks (\textit{PC} and \textit{Dining}). Nonetheless, it mostly fails in the other more complex domains. Since LLM-GenPlan solely utilizes LLMs to propose simple and non-search-based problem-solving strategies, its capability to tackle more complicated problems, i.e., problems with underlying preconditioned sub-ones, is greatly limited.

\textbf{SayPlan} achieves slightly lower success rates in the domains with independent sub-tasks than LLM-GenPlan, but considerably higher success rates in those with dependent sub-tasks ($54\%$ in \textit{Cleaning} and $40\%$ in \textit{Office}), where LLM-GenPlan barely succeeded. Both approaches have replanning mechanisms, but LLM-GenPlan only relies on the plan validator, which checks invalid actions with unsatisfied preconditions. SayPlan on the other hand, obtains feedback from the SG simulator, which additionally provides environmental information when an action fails, e.g., cannot go to an unconnected room, wrong item location, or unaffordable action upon an item. By grounding the actionable knowledge into environmental topology (i.e., SGs), SayPlan is able to tackle more complex tasks then LLM-GenPlan.


\begin{table*}[t]
    \centering
    \renewcommand{\arraystretch}{0.95}
    \setlength{\tabcolsep}{2.2pt}
    \begin{tabular}{c l *{12}{c}}
        \toprule
        \multirow{2}{*}{Metrics} & \multirow{2}{*}{Models} & \multicolumn{3}{c}{PC} & \multicolumn{3}{c}{Dining} & \multicolumn{3}{c}{Cleaning} & \multicolumn{3}{c}{Office} \\
        \cmidrule{3-14} & & A & S & P & A & S & P & A & S & P & A & S & P \\
        \midrule
        \multirow{6}{*}{\makecell{Plan\\Length}} & GT & \textbf{41} & \textbf{42} & \textbf{47} & \textbf{39} & \textbf{39} & \textbf{33} & \textbf{39} & \textbf{43} & \textbf{41} & \textbf{40} & \textbf{33} & \textbf{52} \\
        & LLM-As-Planner           & \textbf{41} & 42.81 & \textbf{47} & - & 43 & 35 & - & - & - & - & - & - \\
        & LLM+P                    & \textbf{41} & \textbf{42} & \textbf{47} & - & \textbf{39} & - & - & - & - & - & - & - \\
        & LLM-GenPlan              & 41.32 & 43.65 & \textbf{47} & 40.95 & 40.70 & 35 & 43.67 & - & 47 & - & 37 & - \\
        & SayPlan                  & 44.45 & 48 & 47.24 & 41.83 & 46.97 & 35.48 & 45 & 48.11 & 43.29 & 46 & 42.85 & 56.47 \\
        & DELTA                    & \textbf{41} & \textbf{42} & \textbf{47} & \textbf{39} & \textbf{39} & 35 & 40 & 44 & 45 & \textbf{40} & \textbf{33} & \textbf{52} \\
        \midrule
        \multirow{2}{*}{\makecell{Planning\\Time}} & DELTA & \textbf{0.0134} & \textbf{0.0144} & \textbf{0.0117} & \textbf{0.0101} & \textbf{0.0103} & \textbf{0.0089} & \textbf{0.0112} & \textbf{0.0120} & \textbf{0.0111} & \textbf{0.0170} & \textbf{0.0167} & \textbf{0.0149} \\
        & DELTA (w/o dp.)      & 51.76 & 49.29 & 28.65 & 42.69 & 54.68 & 1.76 & 23.04 & 58.38 & 5.75 & 24.69 & 9.01 & 10.67 \\
        \midrule
        \multirow{2}{*}{\makecell{Expanded\\Nodes}} & DELTA & \textbf{624.83} & \textbf{727} & \textbf{576} & \textbf{571.83} & \textbf{597.85} & \textbf{561.11} & \textbf{407} & \textbf{364} & \textbf{365.12} & \textbf{405.80} & \textbf{423.03} & \textbf{442.17} \\
        & DELTA (w/o dp.)      & 1,585,185 & 1,317,615 & 1,379,036 & 1,016,429 & 2,187,261 & 368,728 & 1,561,834 & 1,797,434 & 107,400 & 501,148 & 149,024 & 321,221 \\
        \bottomrule
    \end{tabular}
    \caption{Further metrics of DELTA with and without goal decomposition (\textit{w/o dp.}) and other baselines in all domains and A(llensville), S(helbiana), and P(arole) scenes. The ground truth (\textit{GT}) plan lengths are shown in the first row, indicating the optimal values. \textit{``-''} means the metric is not applicable due to failures. All results are based on \textit{GPT-4o} and are averaged over the succeeded cases.\vspace{-1em}}
    \label{tab:result_all}
\end{table*}


Although replanning significantly improves the success rates of LLM-GenPlan and SayPlan, as indicated by their results with \textit{(w/o replan)} in Table \ref{tab:sr}, it merely ensures the executability of the generated plans but not the correctness and optimality, i.e., the plan reaches the goal with the shortest path. This is because LLMs generate output by predicting a probability distribution over the possible next tokens \cite{pallagani2024prospects}, instead of being a search- or an optimization-based planning process. The higher numbers of plan lengths of LLM-GenPlan and SayPlan in Table \ref{tab:result_all} further prove the statement. Thus, LLM-GenPlan and SayPlan are not suitable for long-term task planning, as shown in Table \ref{tab:cap}.

\begin{figure}[t]
    \centering
    \includegraphics[width=0.45\textwidth]{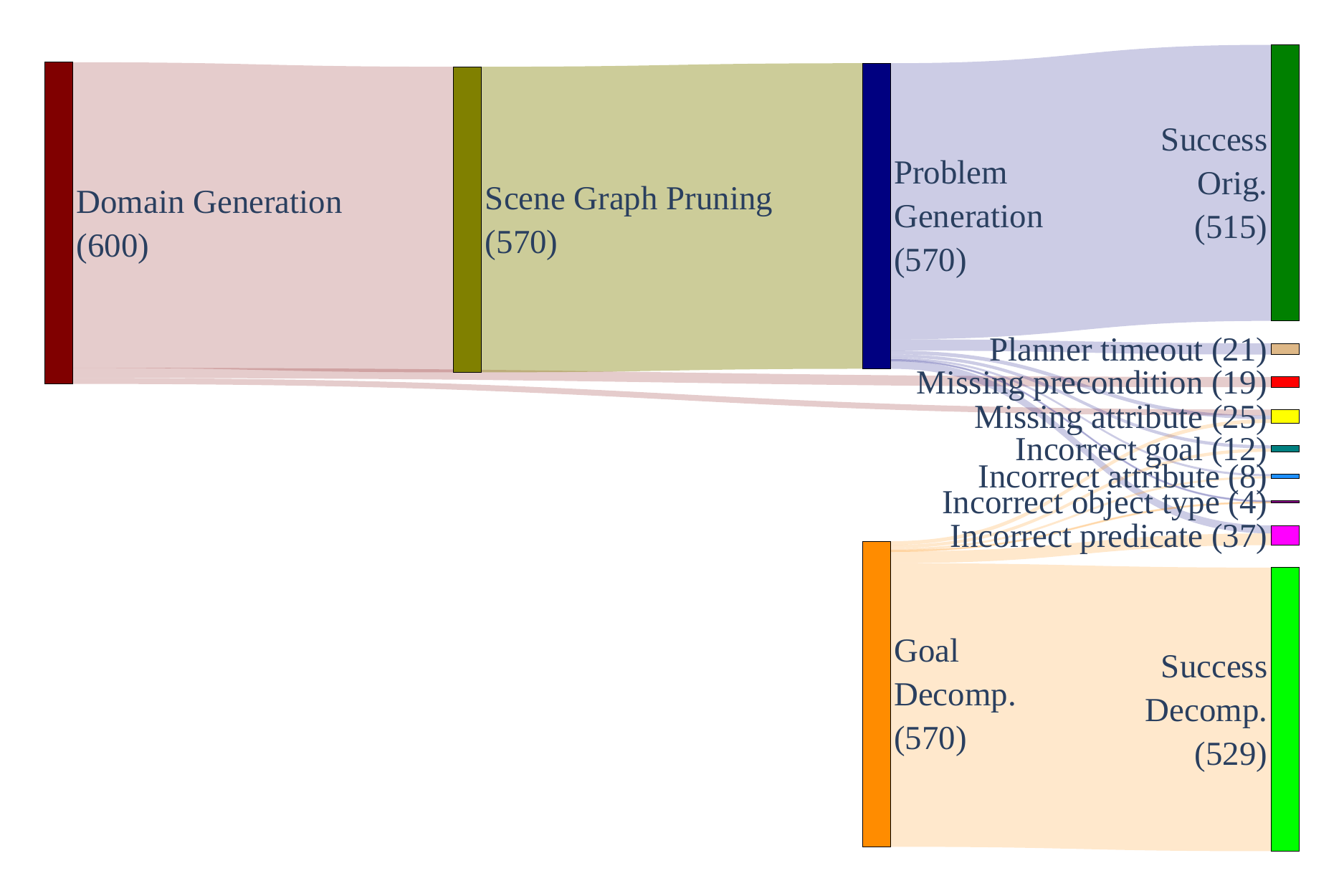}
    \caption{Failure analysis of DELTA with \textit{GPT-4o}. Each step, success, and failure type is annotated with the number of trials. \textit{Problem Generation} and \textit{Goal Decomposition} are decoupled since the planning of original and decomposed problems are independent and executed parallelly with the same number of trials outgoing from \textit{Scene Graph Pruning} step.}
    \label{fig:fail}
\end{figure}

Finally, \textbf{DELTA} achieved the highest success rates in all domains. 
The last two rows of Table \ref{tab:sr} infer that the goal decomposition marginally improves its success rates since the original problems with undecomposed goals have significantly higher complexity, as reflected by the number of expanded nodes in the lower part of Table~\ref{tab:result_all}, which occasionally leads to the planner timeout. Fig. \ref{fig:fail} indicates that $21$ out of the $600$ trials failed due to planner timeout. Further leading causes of failures are incorrectly generated predicates ($37$ out of $600$ trials, such as the \textit{neighbor} relationship of two unconnected rooms), missing attributes ($25$ out of $600$ trials, such as \textit{item\_accessible}), etc.

The key factors that enable DELTA for long-term planning are grounding the actionable knowledge into formal planning language and relying on automated planners to find optimal solutions. 
As indicated in Table \ref{tab:cap} and Sec. \ref{sec:baseline}, with LLM+P being a subset of DELTA focusing solely on problem generation, and both DELTA and LLM+P translating NL into formal language, LLM+P can also handle long-term planning problems thanks to the usage of PDDL. 
However, LLM+P succeeded notably less than DELTA, because apart from goal decomposition, LLM+P also suffers from redundant scene representations, i.e., unpruned SGs, that prevent the automated planner from finding solutions efficiently. 
Having more sparse SGs notably reduces the risk of LLMs producing unnecessary and possibly incorrect predicates translated from the irrelevant items, which results in more potential planning errors.

Decomposing the long-term goal also contributes to a significant reduction of the planning time and the number of expanded nodes by four orders of magnitudes, thus enabling a vast enhancement of planning efficiency. As shown in the lower part of Table \ref{tab:result_all}, the planning time is over $3,000$ times faster in \textit{PC} and \textit{Dining} domains, and almost $2,000$ time faster in \textit{Cleaning} and \textit{Office} domains on average. The number of expanded nodes is also reduced by, on average, circa $2,000$ times in all the experiments.

\textbf{Ablation:} Besides \textit{GPT-4o}, we further evaluate DELTA with \textit{GPT-4-turbo} and \textit{Llama-3.1-70B}. The corresponding success rates are listed in the lower part of Table \ref{tab:sr}. 
Switching to \textit{GPT-4-turbo} results in a notable performance drop, especially in the \textit{Office} domain, where the success rate decreased significantly from $74.67\%$ to $9.66\%$. The results in \textit{PC} and \textit{Dining} domains, i.e., domains with independent sub-tasks, are less affected.
Nonetheless, the numbers from \textit{Llama-3.1-70B} are even considerably lower than those from \textit{GPT-4-turbo}, implying an enormous performance gap. 
This is possibly caused by the reducing parameter count of \textit{Llama-3.1-70B} compared to \textit{GPT-4-turbo} and \textit{GPT-4o}.

\section{CONCLUSION}\label{conclusion}
Despite their training on vast amount of data, LLMs can generate infeasible plans due to hallucinations or missing information. Moreover, classical planning techniques often require extensive annotation or domain-specific heuristics to incorporate context and semantics. To address these challenges and improve plan feasibility and efficiency, we introduced DELTA, a novel LLM-informed task planning approach.
DELTA's integration of scene graphs and LLMs facilitates the rapid generation of precise planning problem descriptions. To enhance planning performance, DELTA decomposes long-term task goals with LLMs into a sequence of sub-goals, enabling automated task planners to efficiently solve complex problems. In our evaluation, we show how DELTA enables a significant enhancement of efficiency in automated task planning in terms of a considerably faster planning time and higher success rates compared to various baselines. 
For future work, we plan to implement repairing mechanisms for handling uncertainties in dynamic environments, and validate our approach on real-world robot operations.


\footnotesize
\bibliographystyle{IEEEtran}
\bibliography{main}

\end{document}

%% file: PDDL.tex
\lstdefinelanguage{PDDL}
{
  sensitive=false,    
  morecomment=[l]{;}, 
  alsoletter={:,-},   
  morekeywords={
    define,domain,problem,not,and,or,when,forall,exists,either,
    :domain,:requirements,:types,:objects,:constants,
    :predicates,:action,:parameters,:precondition,:effect,
    :fluents,:primary-effect,:side-effect,:init,:goal,
    :strips,:adl,:equality,:typing,:conditional-effects,
    :negative-preconditions,:disjunctive-preconditions,
    :existential-preconditions,:universal-preconditions,:quantified-preconditions,
    :functions,assign,increase,decrease,scale-up,scale-down,
    :metric,minimize,maximize,
    :durative-actions,:duration-inequalities,:continuous-effects,
    :durative-action,:duration,:condition
  }
}

%% file: main.bbl
\def\authornoop#1{}
\begin{thebibliography}{10}
\providecommand{\url}[1]{#1}
\csname url@samestyle\endcsname
\providecommand{\newblock}{\relax}
\providecommand{\bibinfo}[2]{#2}
\providecommand{\BIBentrySTDinterwordspacing}{\spaceskip=0pt\relax}
\providecommand{\BIBentryALTinterwordstretchfactor}{4}
\providecommand{\BIBentryALTinterwordspacing}{\spaceskip=\fontdimen2\font plus
\BIBentryALTinterwordstretchfactor\fontdimen3\font minus \fontdimen4\font\relax}
\providecommand{\BIBforeignlanguage}[2]{{%
\expandafter\ifx\csname l@#1\endcsname\relax
\typeout{** WARNING: IEEEtran.bst: No hyphenation pattern has been}%
\typeout{** loaded for the language `#1'. Using the pattern for}%
\typeout{** the default language instead.}%
\else
\language=\csname l@#1\endcsname
\fi
#2}}
\providecommand{\BIBdecl}{\relax}
\BIBdecl

\bibitem{openai2023gpt4}
J.~Achiam \emph{et~al.}, ``Gpt-4 technical report,'' \emph{arXiv preprint arXiv:2303.08774}, 2023.

\bibitem{chowdhery2023palm}
A.~Chowdhery \emph{et~al.}, ``Palm: Scaling language modeling with pathways,'' \emph{Journal of Machine Learning Research}, vol.~24, no. 240, pp. 1--113, 2023.

\bibitem{devlin2018bert}
J.~Devlin, M.-W. Chang, K.~Lee, and K.~Toutanova, ``Bert: Pre-training of deep bidirectional transformers for language understanding,'' \emph{arXiv preprint arXiv:1810.04805}, 2018.

\bibitem{touvron2023llama}
H.~Touvron \emph{et~al.}, ``Llama: Open and efficient foundation language models,'' \emph{arXiv preprint arXiv:2302.13971}, 2023.

\bibitem{aiello2023service}
M.~Aiello and I.~Georgievski, ``Service composition in the chatgpt era,'' \emph{Service Oriented Computing and Applications}, pp. 1--6, 2023.

\bibitem{liu2023human}
Y.~Liu, L.~Palmieri, I.~Georgievski, and M.~Aiello, ``Human-flow-aware long-term mobile robot task planning based on hierarchical reinforcement learning,'' \emph{IEEE Robotics and Automation Letters}, 2023.

\bibitem{aboki2023automating}
N.~Aboki, I.~Georgievski, and M.~Aiello, ``Automating a telepresence robot for human detection, tracking, and following,'' in \emph{Annual Conference Towards Autonomous Robotic Systems}.\hskip 1em plus 0.5em minus 0.4em\relax Springer, 2023.

\bibitem{huang2022language}
W.~Huang, P.~Abbeel, D.~Pathak, and I.~Mordatch, ``Language models as zero-shot planners: Extracting actionable knowledge for embodied agents,'' in \emph{International Conference on Machine Learning}.\hskip 1em plus 0.5em minus 0.4em\relax PMLR, 2022, pp. 9118--9147.

\bibitem{song2023llm}
C.~H. Song, J.~Wu, C.~Washington, B.~M. Sadler, W.-L. Chao, and Y.~Su, ``Llm-planner: Few-shot grounded planning for embodied agents with large language models,'' in \emph{Proceedings of the IEEE/CVF International Conference on Computer Vision}, 2023, pp. 2998--3009.

\bibitem{liu2023llm+}
B.~Liu \emph{et~al.}, ``Llm+ p: Empowering large language models with optimal planning proficiency,'' \emph{arXiv preprint arXiv:2304.11477}, 2023.

\bibitem{ding2023task}
Y.~Ding, X.~Zhang, C.~Paxton, and S.~Zhang, ``Task and motion planning with large language models for object rearrangement,'' \emph{arXiv preprint arXiv:2303.06247}, 2023.

\bibitem{ahn2022i}
M.~Ahn \emph{et~al.}, ``Do as i can, not as i say: Grounding language in robotic affordances,'' \emph{arXiv preprint arXiv:2204.01691}, 2022.

\bibitem{rana2023sayplan}
K.~Rana, J.~Haviland, S.~Garg, J.~Abou-Chakra, I.~Reid, and N.~Suenderhauf, ``Sayplan: Grounding large language models using 3d scene graphs for scalable robot task planning,'' in \emph{Conference on Robot Learning}.\hskip 1em plus 0.5em minus 0.4em\relax PMLR, 2023, pp. 23--72.

\bibitem{chen2023autotamp}
Y.~Chen, J.~Arkin, Y.~Zhang, N.~Roy, and C.~Fan, ``Autotamp: Autoregressive task and motion planning with llms as translators and checkers,'' \emph{arXiv preprint arXiv:2306.06531}, 2023.

\bibitem{silver2024generalized}
T.~Silver, S.~Dan, K.~Srinivas, J.~B. Tenenbaum, L.~Kaelbling, and M.~Katz, ``Generalized planning in {PDDL} domains with pretrained large language models,'' in \emph{Proceedings of the AAAI Conference on Artificial Intelligence}, vol.~38, no.~18, 2024, pp. 20\,256--20\,264.

\bibitem{pallagani2024prospects}
V.~Pallagani \emph{et~al.}, ``On the prospects of incorporating large language models (llms) in automated planning and scheduling (aps),'' in \emph{Proceedings of the International Conference on Automated Planning and Scheduling}, vol.~34, 2024, pp. 432--444.

\bibitem{valmeekam2022large}
K.~Valmeekam, A.~Olmo, S.~Sreedharan, and S.~Kambhampati, ``Large language models still can't plan (a benchmark for llms on planning and reasoning about change),'' \emph{arXiv preprint arXiv:2206.10498}, 2022.

\bibitem{mcdermott1998pddl}
D.~McDermott \emph{et~al.}, ``Pddl-the planning domain definition language,'' 1998.

\bibitem{xie2023translating}
Y.~Xie, C.~Yu, T.~Zhu, J.~Bai, Z.~Gong, and H.~Soh, ``Translating natural language to planning goals with large-language models,'' \emph{arXiv preprint arXiv:2302.05128}, 2023.

\bibitem{zuo2024planetarium}
M.~Zuo, F.~P. Velez, X.~Li, M.~L. Littman, and S.~H. Bach, ``Planetarium: A rigorous benchmark for translating text to structured planning languages,'' \emph{arXiv preprint arXiv:2407.03321}, 2024.

\bibitem{goel2023semantically}
Y.~Goel, N.~Vaskevicius, L.~Palmieri, N.~Chebrolu, K.~O. Arras, and C.~Stachniss, ``Semantically informed mpc for context-aware robot exploration,'' in \emph{2023 IEEE/RSJ International Conference on Intelligent Robots and Systems (IROS)}.\hskip 1em plus 0.5em minus 0.4em\relax IEEE, 2023, pp. 11\,218--11\,225.

\bibitem{chalvatzaki2023learning}
G.~Chalvatzaki, A.~Younes, D.~Nandha, A.~T. Le, L.~F. Ribeiro, and I.~Gurevych, ``Learning to reason over scene graphs: a case study of finetuning gpt-2 into a robot language model for grounded task planning,'' \emph{Frontiers in Robotics and AI}, vol.~10, 2023.

\bibitem{ravichandran2022hierarchical}
Z.~Ravichandran, L.~Peng, N.~Hughes, J.~D. Griffith, and L.~Carlone, ``Hierarchical representations and explicit memory: Learning effective navigation policies on 3d scene graphs using graph neural networks,'' in \emph{2022 International Conference on Robotics and Automation (ICRA)}.\hskip 1em plus 0.5em minus 0.4em\relax IEEE, 2022, pp. 9272--9279.

\bibitem{agia2022taskography}
C.~Agia \emph{et~al.}, ``Taskography: Evaluating robot task planning over large 3d scene graphs,'' in \emph{Conference on Robot Learning}.\hskip 1em plus 0.5em minus 0.4em\relax PMLR, 2022.

\bibitem{Armeni2019hierachical}
I.~Armeni \emph{et~al.}, ``3d scene graph: A structure for unified semantics, 3d space, and camera,'' in \emph{Proceedings of the IEEE/CVF International Conference on Computer Vision (ICCV)}, 2019.

\bibitem{rosinol20203d}
A.~Rosinol, A.~Gupta, M.~Abate, J.~Shi, and L.~Carlone, ``3d dynamic scene graphs: Actionable spatial perception with places, objects, and humans,'' in \emph{Robotics: Science and Systems (RSS)}, 2020.

\bibitem{Rosinol_2021_SAGE}
A.~Rosinol \emph{et~al.}, ``Kimera: From slam to spatial perception with 3d dynamic scene graphs,'' \emph{The International Journal of Robotics Research}, vol.~40, no. 12-14, pp. 1510--1546, 2021.

\bibitem{hughes2022hydra}
N.~Hughes, Y.~Chang, and L.~Carlone, ``Hydra: A real-time spatial perception system for {3D} scene graph construction and optimization,'' 2022.

\bibitem{Wald2020D3SSG}
J.~Wald, H.~Dhamo, N.~Navab, and F.~Tombari, ``Learning 3d semantic scene graphs from 3d indoor reconstructions,'' in \emph{Proceedings of the IEEE/CVF Conference on Computer Vision and Pattern Recognition (CVPR)}, 2020.

\bibitem{Wu_2021_sgfusion}
S.-C. Wu, J.~Wald, K.~Tateno, N.~Navab, and F.~Tombari, ``Scenegraphfusion: Incremental 3d scene graph prediction from rgb-d sequences,'' in \emph{Proceedings of the IEEE/CVF Conference on Computer Vision and Pattern Recognition (CVPR)}, June 2021, pp. 7515--7525.

\bibitem{Wang_2023_vlsat}
Z.~Wang, B.~Cheng, L.~Zhao, D.~Xu, Y.~Tang, and L.~Sheng, ``Vl-sat: Visual-linguistic semantics assisted training for 3d semantic scene graph prediction in point cloud,'' in \emph{Proceedings of the IEEE/CVF Conference on Computer Vision and Pattern Recognition (CVPR)}, June 2023, pp. 21\,560--21\,569.

\bibitem{koch2024sgrec}
S.~Koch, P.~Hermosilla, N.~Vaskevicius, M.~Colosi, and T.~Ropinski, ``Sgrec3d: Self-supervised 3d scene graph learning via object-level scene reconstruction,'' in \emph{Proceedings of the IEEE/CVF Winter Conference on Applications of Computer Vision (WACV)}, 2024.

\bibitem{koch2023lang3dsg}
S.~{\authornoop{K}}Koch, P.~Hermosilla, N.~Vaskevicius, M.~Colosi, and T.~Ropinski, ``Lang3dsg: Language-based contrastive pre-training for 3d scene graph prediction,'' \emph{arXiv preprint arXiv:2310.16494}, 2023.

\bibitem{conceptgraphs}
Q.~Gu \emph{et~al.}, ``Conceptgraphs: Open-vocabulary 3d scene graphs for perception and planning,'' \emph{arXiv preprint arXiv:2309.16650}, 2023.

\bibitem{chang2023contextaware}
H.~Chang \emph{et~al.}, ``Context-aware entity grounding with open-vocabulary 3d scene graphs,'' in \emph{7th Annual Conference on Robot Learning}, 2023.

\bibitem{koch2024open3dsg}
S.~Koch, N.~Vaskevicius, M.~Colosi, P.~Hermosilla, and T.~Ropinski, ``Open3dsg: Open-vocabulary 3d scene graphs from point clouds with queryable objects and open-set relationships,'' \emph{arXiv preprint arXiv:2402.12259}, 2024.

\bibitem{seymour2022graphmapper}
Z.~Seymour, N.~C. Mithun, H.-P. Chiu, S.~Samarasekera, and R.~Kumar, ``Graphmapper: Efficient visual navigation by scene graph generation,'' in \emph{2022 26th International Conference on Pattern Recognition (ICPR)}.\hskip 1em plus 0.5em minus 0.4em\relax IEEE, 2022, pp. 4146--4153.

\bibitem{xiong2024large}
S.~Xiong, A.~Payani, R.~Kompella, and F.~Fekri, ``Large language models can learn temporal reasoning,'' \emph{arXiv preprint arXiv:2401.06853}, 2024.

\bibitem{yang2023harnessing}
Y.~Yang, S.~Xiong, A.~Payani, E.~Shareghi, and F.~Fekri, ``Harnessing the power of large language models for natural language to first-order logic translation,'' \emph{arXiv preprint arXiv:2305.15541}, 2023.

\bibitem{brown2020language}
T.~Brown \emph{et~al.}, ``Language models are few-shot learners,'' \emph{Advances in neural information processing systems}, vol.~33, pp. 1877--1901, 2020.

\bibitem{chen2023open}
B.~Chen \emph{et~al.}, ``Open-vocabulary queryable scene representations for real world planning,'' in \emph{2023 IEEE International Conference on Robotics and Automation (ICRA)}.\hskip 1em plus 0.5em minus 0.4em\relax IEEE, 2023, pp. 11\,509--11\,522.

\bibitem{shah2023lm}
D.~Shah, B.~Osi{\'n}ski, S.~Levine \emph{et~al.}, ``Lm-nav: Robotic navigation with large pre-trained models of language, vision, and action,'' in \emph{Conference on Robot Learning}.\hskip 1em plus 0.5em minus 0.4em\relax PMLR, 2023, pp. 492--504.

\bibitem{wang2022less}
S.~Wang \emph{et~al.}, ``Less is more: Generating grounded navigation instructions from landmarks,'' in \emph{Proceedings of the IEEE/CVF Conference on Computer Vision and Pattern Recognition}, 2022, pp. 15\,428--15\,438.

\bibitem{helmert2006fast}
M.~Helmert, ``{The Fast Downward Planning System},'' \emph{Journal of Artificial Intelligence Research}, vol.~26, pp. 191--246, 2006.

\bibitem{silver2020pddlgym}
T.~Silver and R.~Chitnis, ``Pddlgym: Gym environments from pddl problems,'' in \emph{International Conference on Automated Planning and Scheduling (ICAPS) PRL Workshop}, 2020.

\bibitem{howey2004val}
R.~Howey, D.~Long, and M.~Fox, ``{VAL: automatic plan validation, continuous effects and mixed initiative planning using PDDL},'' in \emph{16th IEEE International Conference on Tools with Artificial Intelligence}.\hskip 1em plus 0.5em minus 0.4em\relax IEEE, 2004, pp. 294--301.

\end{thebibliography}
